\documentclass[10pt,a4paper,english]{article}
\usepackage{abstract}
\usepackage{multicol}

\usepackage{tipa}

\usepackage[left=2.7cm,right=2.7cm,top=2cm,bottom=2cm]{geometry}
\usepackage[utf8]{inputenc}
\usepackage[USenglish]{babel}
\usepackage[T1]{fontenc}
\usepackage{csquotes}
\usepackage{float}
\usepackage[bottom]{footmisc} 
\usepackage{setspace} 
\usepackage{ragged2e} 
\usepackage[parfill]{parskip} 
\usepackage{pdfpages} 

\usepackage{subfigure}
\usepackage{booktabs}

\usepackage{xcolor}
\definecolor{heidelberg-red}{HTML}{E50A37}
\usepackage{graphicx}
\usepackage{caption}

\usepackage{amsmath}
\usepackage{amsfonts}
\usepackage{amsthm}
\usepackage{amssymb}
\usepackage{bm}
\usepackage{cancel}

\usepackage{witharrows} 
\usepackage{nicefrac} 
\usepackage{physics}
\usepackage{siunitx}
\usepackage{empheq} 

\usepackage[automake]{glossaries}
\setacronymstyle{long-short}

\usepackage{lmodern}
\usepackage{fancyhdr}
\usepackage{enumerate}
\usepackage{hyperref} 
\hypersetup{
	colorlinks = true,
	linkcolor={red!50!black}, 
	citecolor={blue!50!black}, 
	urlcolor={blue!80!black} 
}
\usepackage{todonotes}
\usepackage[nameinlink]{cleveref}

\let\qty\SI 

\usepackage{enumitem}
\setenumerate[0]{label=\textbf{\alph*)}}
\setlist{leftmargin=*}

\usepackage{float}
\usepackage[ruled, algosection, lined, scleft, linesnumbered, noend]{algorithm2e}

\SetCommentSty{mycommentsfont} 
\SetKwComment{Comment}{$\triangleright$\ }{} 
\SetKwRepeat{DoWhile}{do}{while}

\SetKwProg{Fn}{Function}{:}{}
\SetArgSty{textnormal}

\SetKw{Continue}{continue} 
\SetKw{Break}{break}

\usepackage[most]{tcolorbox}
\tcbset{
    frame empty,
    colback=lightgray!20
}

\usepackage[
    backend=biber,
    style=numeric,
    sorting=none
]{biblatex}

\newcommand{\eg}{e.\,g. }
\newcommand{\ie}{i.\,e. }

\newcommand{\q}[1]{\enquote{#1}}

\newcommand{\set}[1]{\ensuremath{\{\,#1\,\}}}
\renewcommand{\implies}{\Rightarrow}
\renewcommand{\iff}{\Leftrightarrow}

\newcommand\Cpp{C\nolinebreak[4]\hspace{-.05em}\raisebox{.4ex}{\relsize{-3}{\textbf{++}}}}

\makeatletter
\newlength\xvec@height%
\newlength\xvec@depth%
\newlength\xvec@width%
\newcommand{\xvec}[2][]{%
	\ifmmode%
	\settoheight{\xvec@height}{$#2$}%
	\settodepth{\xvec@depth}{$#2$}%
	\settowidth{\xvec@width}{$#2$}%
	\else%
	\settoheight{\xvec@height}{#2}%
	\settodepth{\xvec@depth}{#2}%
	\settowidth{\xvec@width}{#2}%
	\fi%
	\def\xvec@arg{#1}%
	\def\xvec@dd{:}%
	\def\xvec@d{.}%
	\raisebox{.2ex}{\raisebox{\xvec@height}{\rlap{%
				\kern.05em
				\begin{tikzpicture}[scale=1]
					\pgfsetroundcap
					\draw (.05em,0)--(\xvec@width-.05em,0);
					\draw (\xvec@width-.05em,0)--(\xvec@width-.15em, .075em);
					\draw (\xvec@width-.05em,0)--(\xvec@width-.15em,-.075em);
					\ifx\xvec@arg\xvec@d%
					\fill(\xvec@width*.45,.5ex) circle (.5pt);%
					\else\ifx\xvec@arg\xvec@dd%
					\fill(\xvec@width*.30,.5ex) circle (.5pt);%
					\fill(\xvec@width*.65,.5ex) circle (.5pt);%
					\fi\fi%
				\end{tikzpicture}%
	}}}%
	#2%
}
\makeatother

\title{\vspace{-0.0em}Parallel Needleman–Wunsch on CUDA to measure\\
word similarity based on phonetic transcriptions}
\author{Dominic Plein}
\date{September 1, 2025}

\newcommand{\abstractText}{\noindent
	\newline\noindent
    We present a method to calculate the similarity between words based on their phonetic transcription (their pronunciation) using the Needleman–Wunsch algorithm. We implement this algorithm in Rust and parallelize it on both CPU and GPU to handle large datasets efficiently. The GPU implementation leverages CUDA and the cudarc Rust library to achieve significant performance improvements. We validate our approach by constructing a fully-connected graph where nodes represent words and edges have weights according to the similarity between the words. This graph is then analyzed using clustering algorithms to identify groups of phonetically similar words. Our results demonstrate the feasibility and effectiveness of the proposed method in analyzing the phonetic structure of languages. It might be easily expanded to other languages.
}

\makeglossaries
\newacronym{ipa}{IPA}{International Phonetic Alphabet}
\newacronym{pos}{POS}{Part-of-Speech}
\newacronym{ptx}{PTX}{Parallel Thread Execution}

\begin{document}

\setlength{\abovedisplayskip}{0.2em}

\maketitle

\begin{abstract}
    \abstractText
    \newline
    \newline
\end{abstract}

\vspace{-2em}
\begin{center}
    This paper is accompanied by a \href{https://youtu.be/xbcpnItE3_4}{YouTube video} and a \href{https://github.com/Splines/phonetics-graph/}{GitHub repository}.
\end{center}

\vspace{2em}

\begin{multicols*}{2}
\tableofcontents
\section{Introduction}

The \gls{ipa} uses special symbols\footnote{See for example the French list \href{https://en.wikipedia.org/wiki/Help:IPA/French}{here}.} to represent the sound of a spoken language \cite{ipa}. This is useful for language learners since the pronunciation of a word can be significantly different from its written form. For example, the French word \textit{renseignement} (information) is pronounced \textipa{/K\~{a}.sE\textltailn.m\~{a}/}. Based on this alphabet, one might wonder if we can construct a metric that quantifies the \textbf{similarity between two words based on their phonetic transcription}. This would allow to construct a graph where nodes are words and undirected edges are weighted by the distance between the words. Such a graph can be used to find neighbors of a word based on the respective phonetic similarity. This opens up the possibility to apply clustering algorithms and other methods stemming from graph theory in order to analyze the phonetic structure of a language.

Calculating the distance between each pair of words corresponds to a fully-connected graph. Our dataset consists of around 600,000 French words and their \gls{ipa} transcription, alongside their frequency in the French language. Excluding self-loops, we find a vast number of edges \eqref{eq:num-edges}:
\begin{align}
    \text{\#nodes} &= 600,000 \\
    \text{\#edges} &= \frac{600,000 \cdot 599,999}{2} \approx \num{1.8e11}
\end{align}
This high number and the independent nature of the distance calculation for each pair of words makes the problem well-suited for parallelization. In \autoref{sec:needleman-wunsch}, we present the Needleman–Wunsch algorithm used to calculate the distance between two words. In \autoref{sec:impl}, we discuss how to parallelize this algorithm on a CPU using the \textit{Rayon} library in Rust and on a consumer Nvidia GPU using the CUDA framework with the \textit{cudarc} Rust library. Finally, we present performance results and provide visualizations of the obtained graphs in \autoref{sec:eval}. We conclude in \autoref{sec:conclusion} and discuss how this method can be improved and how future work could extend it.

\newcolumn
\section{Needleman–Wunsch}
\label{sec:needleman-wunsch}
\newcommand{\lenn}{\text{len}}

The Needleman–Wunsch algorithm calculates the global alignment of two strings and was originally used in bio-informatics to compare amino acid sequences of two proteins \cite{nw}. For our purposes, the alphabet will instead consist of the phonetic \gls{ipa} symbols. Out of all possible alignments of two words (including gaps), the Needleman–Wunsch algorithm finds the one with the smallest distance, \ie the alignment with the highest \q{score}. The algorithm is based on dynamic programming and has a time complexity of $\mathcal{O}(\lenn(A) \cdot \lenn(B))$, where $A$ and $B$ are the two words to be compared.

\begin{figure}[H]
    \centering
    \includegraphics[width=0.77\linewidth]{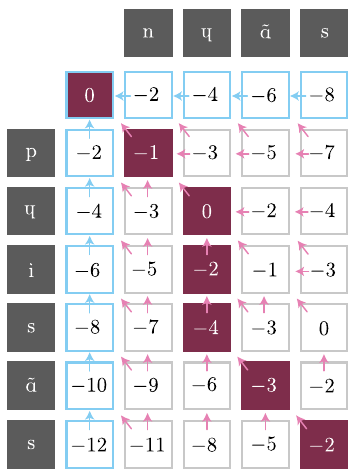}
    \caption{Needleman–Wunsch score matrix for the words $A\coloneqq\textit{puissance}$ \textipa{/p\textturnh is\~As/} (power, strength) and $B\coloneqq\textit{nuance}$ \textipa{/n\textturnh\~As/} (nuance, shade). The arrows indicate which steps locally maximize the score. The red tiles trace the path of the optimal alignment. Match Score: $1$, Mismatch Score: $-1$, Gap Penalty: $p=-2$.}
    \label{fig:matrix-nuance-puissance-optimal}
\end{figure}

\autoref{alg:needleman-wunsch} features the pseudo-code of the score computation\footnote{The respective \href{https://en.wikipedia.org/wiki/Needleman\%E2\%80\%93Wunsch_algorithm}{Wikipedia page} also provides a good introduction. Furthermore, the score matrix is interactively explained in \cite{nw_demo}.}. In \autoref{fig:matrix-nuance-puissance-optimal}, we see the resulting score matrix that the algorithm constructed for the French words \textit{puissance} and \textit{nuance}. Follow the indicated path (red tiles) from the bottom right to the top left to find the (reversed) optimal alignment (see \autoref{tab:nuance-puissance-alignment-optimal}).

\begin{table}[H]
\centering
\begin{tabular}{l*{6}{>{\centering\arraybackslash}p{0.2cm}}}
    \toprule
    \textit{puissance}
    & \textipa{p} & \textipa{\textturnh} & \textipa{i} & \textipa{s} & \textipa{\~A} & \textipa{s}\\
    \midrule
    \textit{nuance}
    & \textipa{n} & \textipa{\textturnh} & -- & -- & \textipa{\~A} & \textipa{s}\\
    \bottomrule
\end{tabular}
\caption{The optimal alignment yields a score of $-2$. See the path in \autoref{fig:matrix-nuance-puissance-optimal}.}
\label{tab:nuance-puissance-alignment-optimal}
\end{table}

\vfill\null

We first discuss the meaning of the different steps (arrows) in the score matrix (\autoref{fig:matrix-nuance-puissance-optimal}) to then explain how to construct this matrix.

\begin{itemize}

    \item In a \textbf{diagonal step}, both symbols (which indicate the current position in the two words) change. Such a step corresponds to either a match or mismatch between the two symbols. In the example, the \textipa{/s/} symbols in the bottom-right corner match, which is why the step beforehand is a \textit{diagonal} step from the field $-3$ to $-2$. The score increases by $1$ since we defined the match score to be $+1$.

    \item In a \textbf{vertical} or \textbf{horizontal step}, only one of the two symbols changes. We interpret this as a gap in the alignment, \ie one symbol aligns to a gap in the other word. In the example, this is the case two times when we move from the red field $0$ down to $-2$ and then down to $-4$. The score decreases by $2$ each time, as we defined the gap penalty as $p \coloneqq -2$ in this example. The gap is indicated by \q{--} in the alignment (see \autoref{tab:nuance-puissance-alignment-optimal}). As we are still in the column of \textipa{/\textturnh/} of the word \textipa{/n\textturnh\~As/}, we insert two \q{--} symbols after the \textipa{/\textturnh/}. This step is sometimes also referred to as \textbf{deletion} or \textbf{insertion}.
    
\end{itemize}

To find the score matrix for given input words $A$ and $B$, we follow \autoref{alg:needleman-wunsch}. First, the score matrix of dimension $(\lenn(A)+1) \times (\lenn(B)+1)$ is initialized\footnote{This does not necessarily involve setting all fields to $0$ as will become clear.}. Then, in lines~\ref{algstep:init-gap-start} to~\ref{algstep:init-gap-end}, the blue-bordered tiles of \autoref{fig:matrix-nuance-puissance-optimal} are filled with the gap penalty $p$ times the index. This is necessary since the only possible step for these tiles is either a vertical or horizontal step (blue arrows), thus leading to a gap in the alignment as discussed beforehand. This gap is punished with gap penalty $p$.

\begin{figure}[H]
    \centering
    \includegraphics[width=0.6\linewidth]{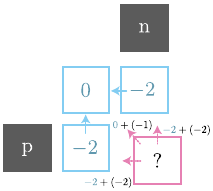}
    \caption{Calculations for one element of the Needleman–Wunsch score matrix.}
    \label{fig:matrix-nuance-puissance-subcalc}
\end{figure}

In the nested loops (lines~\ref{algstep:nested1} and~\ref{algstep:nested2}), we then iterate over the remaining fields of the score matrix (index now starts at $1$, not $0$) which corresponds to traversing the matrix row-wise. To each field, we assign the maximum of three values:

\begin{figure*}

\begin{minipage}[t]{0.6\textwidth}

\begin{algorithm}[H]
    \DontPrintSemicolon
    
    \SetKwFunction{calcScoreFunc}{calculateScore}
    \SetKwData{WordA}{$A$}
    \SetKwData{WordB}{$B$}
    \SetKwData{Sim}{similarity}
    \SetKwData{GapPenalty}{$p$}
    \SetKwData{ScoreMatrix}{scoreMatrix}
    \newcommand{\ScoreMatrixIdx}[2]{{\ScoreMatrix}[{#1}][{#2}]}
    \SetKwData{Cost}{cost}
    \SetKwData{MatchScore}{matchScore}
    \SetKwData{DeleteScore}{deleteScore}
    \SetKwData{InsertScore}{insertScore}
    \SetKwData{Score}{score}
    \SetKwFunction{len}{len}

    \KwIn{
        \small{\!\!\small{$\WordA = \set{A_0, \ldots, A_{\len(A)-1}}$,
        $\WordB = \set{B_0, \ldots, B_{\len(B)-1}}$}},\\
        \qquad\qquad \Sim: \small{similarityScoreFunc},
        \GapPenalty: GapPenalty
    }
    \KwOut{\Score}
    \Fn{\calcScoreFunc{}}
    {
        \SetInd{0.25em}{0.55em}
        Init \ScoreMatrix with dimensions $\small{(\len(\WordA)+1) \times (\len(\WordB)+1)}$\;

        \BlankLine

        \For{$i \in \{0, \ldots, \len(\WordA)\}$\label{algstep:init-gap-start}}
        {
            \ScoreMatrixIdx{$i$}{$0$} $\gets \GapPenalty \cdot i$\;
        }
        \For{$j \in \{0, \ldots, \len(\WordB)\}$}
        {
            \ScoreMatrixIdx{$0$}{$j$} $\gets \GapPenalty \cdot j$
            \label{algstep:init-gap-end}\;
        }

        \BlankLine

        \For{$i \in \{1, \ldots, \len(\WordA)\}$\label{algstep:nested1}}
        {
            \For{$j \in \{1, \ldots, \len(\WordB)\}$\label{algstep:nested2}}
            {
                \Cost $\gets$ $\Sim(\WordA_{i-1}, \WordB_{j-1})$
                \label{algstep:sim}\;
                \MatchScore $\gets$ \ScoreMatrixIdx{$i-1$}{$j-1$} + \Cost
                \label{algstep:matchscore}\;
                \DeleteScore $\gets$ \ScoreMatrixIdx{$i-1$}{$j$} + \GapPenalty
                \label{algstep:deletescore}\;
                \InsertScore $\gets$ \ScoreMatrixIdx{$i$}{$j-1$} + \GapPenalty
                \label{algstep:insertscore}\;
                \ScoreMatrixIdx{$i$}{$j$} $\gets$
                $\max(\MatchScore, \DeleteScore, \InsertScore)$
                \label{algstep:max}\;
            }
        }

        \BlankLine

        \Return\! \ScoreMatrixIdx{$\len(\WordA)$}{$\len(\WordB)$}\;
    }
    
    \caption{Needleman–Wunsch}
    \label{alg:needleman-wunsch}
\end{algorithm}

\end{minipage}%
\hfill
\begin{minipage}[t]{0.35\textwidth}

\begin{figure}[H]
    \centering
    \includegraphics[width=0.87\linewidth]{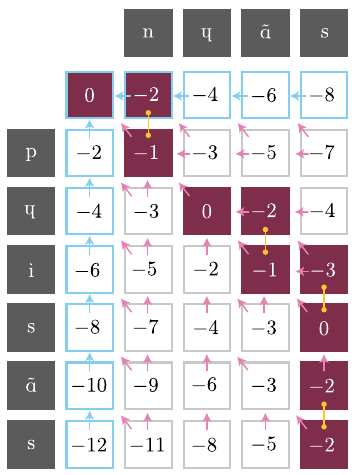}
    \caption{Needleman–Wunsch score matrix and the path (in red) for a non-optimal alignment. Orange strokes indicate non-optimal steps. Parameters as in \autoref{fig:matrix-nuance-puissance-optimal}.}
    \label{fig:matrix-nuance-puissance-non-optimal}
\end{figure}

\end{minipage}
\end{figure*}

\begin{itemize}[leftmargin=0cm]

    \item The \textbf{match score} is calculated by checking the step to the upper left diagonal (\autoref{algstep:matchscore}). In the example of \autoref{fig:matrix-nuance-puissance-subcalc}, this would result in a value $0+(-1) = -1$, where $0$ is the value in the upper left diagonal field and $-1$ is the cost of the mismatch between \textipa{/p/} and \textipa{/n/}. In case of a match, the new value would be $0 + 1 = 1$. In the algorithm, we also consider the case where costs for a match and a mismatch depend on the symbols themselves, which is why we introduce the function \textit{similarity} that returns the cost of aligning two symbols. This is especially useful when comparing phonetic symbols, as the similarity between two symbols can be defined in a more sophisticated way than just $1$ or $-1$ (\eg replacing a vowel with a consonant might be more costly than replacing a vowel with another vowel). This function might also be represented as a \q{similarity matrix}.
    
    \item The \textbf{delete score} refers to the step from the field above (\autoref{algstep:deletescore}). In the example, we find $(-2) + (-2) = -4$ as new value ($-2$ is the value in the field above and $p=-2$ is the gap penalty). This steps signifies that a symbol in word A aligns to a gap in word B (here: \textipa{/i/} and \textipa{/s/} of \textit{puissance} align to gaps in \textit{nuance}).
    
    \item The \textbf{insert score} refers to the step from the left (\autoref{algstep:insertscore}). In the example, we find $(-2) + (-2) = -4$ as new value ($-2$ is the value in the field to the left and $p=-2$ is the gap penalty). This step signifies that a symbol in word B aligns to a gap in word A (this does not occur in the example).

\end{itemize}

The new value of the current field is assigned the maximum of these values (\autoref{algstep:max}), such that we locally maximize the score: $\max(-1, -4, -4) = -1$. In \autoref{fig:matrix-nuance-puissance-optimal}, we additionally kept track of the steps that led to the optimal alignment by means of the rose arrows (sometimes multiple optimal steps are possible). For our purposes, we don't want to reconstruct the exact alignment that led to the optimal score, but only the score itself. Thus, we can omit the backtracking step and don't need to store the rose arrows.

By construction, \textbf{the bottom-right field of the score matrix contains the score of the optimal alignment}. This is ensured by the Principle of Optimality \cite{dp}, which states that an optimal solution to a problem can be constructed from optimal solutions to its subproblems. In the context of the Needleman–Wunsch algorithm, this means that the optimal alignment score for two sequences (words) can be derived by considering the optimal alignment scores of progressively smaller subsequences. Each cell in the score matrix represents the optimal score for the corresponding prefixes of the two words up to that point, since we take the maximum of the three possible steps (match, delete, insert) at each cell. This ensures that the final cell (in the bottom right) contains the optimal score for the entire sequences.

\autoref{tab:nuance-puissance-alignment-non-optimal} shows an example of a non-optimal alignment of the two words, yielding a score of $-15$ (compared to $-2$ for the optimal path). \autoref{fig:matrix-nuance-puissance-non-optimal} depicts the corresponding score matrix. Note how the indicated path includes 4 non-optimal choices (orange strokes).

\vspace{-0.65em}

\begin{table}[H]
    \centering
    \begin{tabular}{l*{9}{>{\centering\arraybackslash}p{0.2cm}}}
        \toprule
        \textit{puissance}
        & -- & \textipa{p} & \textipa{\textturnh} & -- & \textipa{i}
        & -- & \textipa{s} & \textipa{\~A} & \textipa{s}\\
        \midrule
        \textit{nuance}
        & \textipa{n} & -- & \textipa{\textturnh} & \textipa{\~A} & --
        & \textipa{s} & -- & -- & --\\
        \bottomrule
    \end{tabular}
    \caption{This non-optimal alignment yields a score of $-15$. See the path in \autoref{fig:matrix-nuance-puissance-non-optimal}.}
    \label{tab:nuance-puissance-alignment-non-optimal}
\end{table}

\begin{figure*}
    \centering
    \includegraphics[width=0.7\textwidth]{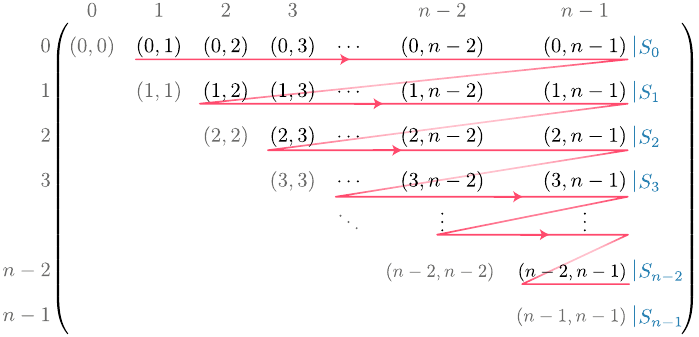}
    \caption{Row-major traversal of the adjacency matrix. $n$ is the total number of words (\ie nodes).}
    \label{fig:traverse-schema}
\end{figure*}

\section{Data Preparation}
\label{sec:data}

We use the \href{https://github.com/frodonh/french-words}{\textbf{french-words}} dataset \cite{data_french_words}, which contains 691,969 French words. The limitation to the French language is arbitrary; the method presented here can be applied to any language that has phonetic transcriptions available, simply by replacing the underlying dataset.

It is compiled from several sources including (among others) the Debian package \href{https://packages.debian.org/fr/sid/wfrench}{wfrench}, \href{http://www.lexique.org/}{Lexique 3.83} \cite{data_lexique}, the \href{https://infolingu.univ-mlv.fr/DonneesLinguistiques/Dictionnaires/telechargement.html}{DELA dictionary} \cite{data_dela} as well as the \href{https://github.com/hbenbel/French-Dictionary}{French-Dictionary} \cite{data_french_csv}. The words also contain \acrfull{pos} tagging information, \eg whether it is a noun, verb, adjective etc. It also comprises the usage frequency according to Lexique.org and Google Ngrams\footnote{We use the average of both sources (or just one if the other is missing).}.

Since the french-words dataset does not include phonetic transcriptions, we merge it with the \href{https://github.com/DanielSWolf/wiki-pronunciation-dict}{\textbf{wiki-pronunciation-dict}} \cite{data_pronunciation} extracted from the French \href{https://fr.wiktionary.org/}{Wiktionnaire} to obtain 611,786 words with their \gls{ipa} transcription\footnote{If multiple transcriptions for one word are available, we (arbitrarily) only store the first one to ease data handling.}. We validate the merged dataset by manually checking random samples of the words. While the dataset is of high quality, it does occasionnally contain errors in the transcriptions that we will not address here.

For usage in the Needleman–Wunsch algorithm, we extract all used phonetic \acrshort{ipa} symbols and assign integer IDs to them. For our examples, we obtain \autoref{tab:phonetic-encoding}. Note that we consider \textipa{/dZ/} as one symbol, even though it is a combination of \textipa{/d/} and \textipa{/Z/}. The same applies to \textipa{/tS/}. This is to account for the different pronunciation of the combined symbols compared to the individual ones.

\begin{table}[H]
    \centering
    \begin{tabular}{lll}
    \toprule
    \textbf{Word} & \textbf{\acrshort{ipa}} & \textbf{Encoding} \\
    \midrule
    \textit{puissance} & \textipa{/p\textturnh is\~As/} & $[0,18,16,11,26,11]$ \\
    \textit{nuance} & \textipa{/n\textturnh\~As/} & $[29,18,26,11]$ \\
    \bottomrule
    \end{tabular}
    \caption{Example of two words with their phonetic transcription and encoding.}
    \label{tab:phonetic-encoding}
\end{table}

\section{Parallelized algorithms}
\label{sec:impl}

In the following, for simplicity we always use a similarity matrix with $1$ on its diagonal (symbols match) and $-1$ elsewhere (symbols do not match). The gap penalty is set to $p=-1$ (instead of $-2$ beforehand in the examples in \autoref{sec:needleman-wunsch}).

\subsection{On the CPU}

A first and slow Python implementation serves as a reference to confirm correctness of the Rust implementation. The latter is parallelized on the CPU by employing the \tcboxverb{rayon} library on the outer for-loop (\autoref{algstep:nested1}): for every word $A$, we consider all other words $B$ and calculate the similarity between $A$ and $B$. Since words lengths differ, the core calculation takes different times for different word pairs. Therefore, every thread appends its result to a vector that is wrapped in a \tcboxverb{Mutex} (avoid data races by mutual exclusion) and an \tcboxverb{Arc} (thread-safe reference pointer to deallocate data at the end).

Since comparing $A$ to $B$ yields the same score as the comparison of $B$ to $A$, we deal with \textit{undirected} edges and thus the adjacency matrix is symmetric. Furthermore, we are not interested in self-loops, but only in the similarity between \textit{different} words. For these two reasons, \textbf{we only consider the upper triangular part of the adjacency matrix} in \autoref{fig:traverse-schema}.

To store the resulting edge weights in a binary file, we traverse the matrix in row-major order (red path): $(0,1)$, $(0,2)$, \ldots, $(0,n-1)$, $(1,2)$, \ldots, $(1,n-1), (2,3), \ldots, (n-2,n-1)$, where $n$ is the total number of words. Similarity scores are encoded as 8-bit signed integers (range $[-128,127]$). This is sufficient since word lengths are typically small and match/mismatch score as well as gap penalty $p$ are likewise chosen to be small.

\subsection{On the GPU}

We implement the algorithm in the CUDA framework and consult the \tcboxverb{cudarc} Rust library supplying Rust wrappers around the CUDA driver API as well as the NVRTC API (among others). The latter makes available methods to compile our \Cpp~kernel to \gls{ptx} code during runtime and to launch it.

\textbf{Subtasks.} Foster's PCAM methodology \cite{foster} can help in designing parallel algorithms. The first step is to partition the problem at hand into small tasks. At the level of the adjacency matrix (\autoref{fig:traverse-schema}), such a task would be to compute the similarity score between two words $A$ and $B$ at the respective row and column. On a finer granularity, we can also refer to \autoref{fig:matrix-nuance-puissance-subcalc} and consider the calculation of one element of the score matrix. However, we realize that latter calculations are highly dependent on each other since the score of a cell depends on the scores of its up, left and up-left (diagonal) neighbors (example of a \textit{stencil computation}). Additionally, words length differ, so the size of the score matrix varies. Due to the data dependencies and the varying problem size, we decide to focus on parallelizing the for-loops in the Needleman–Wunsch algorithm (lines \ref{algstep:nested1} and \ref{algstep:nested2} in \autoref{alg:needleman-wunsch}) making this a \textit{pleasingly parallel} problem, namely that of parallelizing score calculation for entries in the adjacency matrix.

\begin{figure}[H]
    \centering
    \includegraphics[width=0.45\linewidth]{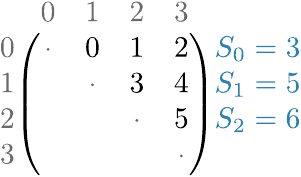}
    \caption{Example of \text{idx} (black) for $n=4$. $S_r$ is the number of elements traversed up to the end of row $r$ (blue).}
    \label{fig:sum-example}
\end{figure}

\textbf{Grid.} The maximum number of blocks in a grid (grid size) is limited to $2^{16} - 1 = 65535$ in $y$ and $z$ direction. Starting with CUDA compute capability 3.0, the limit in the $x$ direction is raised to $2^{31} - 1 = 2,147,483,647$ blocks. Furthermore, we only consider the upper triangular part of the adjacency matrix, while a grid is always rectangular. These constraints motivate the choice of using a one-dimensional grid and spawning blocks only in the $x$ direction. Likewise, we only employ one-dimensional blocks, \ie a block is a one-dimensional array of threads. The exact number of threads we use inside a block (block dimension) is discussed later. With this configuration, we calculate (in every thread) an \textbf{index} ranging from $0$ to a number greater than that of entries in the upper triangular part of the adjacency matrix:
\begin{align}
    \text{idx} = \text{blockIdx.x} \cdot \text{blockDim.x} + \text{threadIdx.x}
    \label{eq:idx}
\end{align}
\textbf{Row \& Column}. The index \tcboxverb{idx} needs to be mapped to the corresponding row and column in the adjacency matrix, such that a thread knows which words to compare. We take advantage of the row-major traversal path. Let $S_r$ be the number of elements traversed up to the end of row $r$. In \autoref{fig:traverse-schema}, this variable is drawn in blue. \autoref{fig:sum-example} provides an example to ease keeping track during following calculations.

\vspace{-3em}

\begin{align}
    S_0 &= n-1, \quad S_1 = S_0 + (n-2), \quad\ldots\\
    S_r &= \sum_{k=0}^{r} (n-k-1)\\[-10pt]
    &= (n-1) \cdot (r+1) - \sum_{k=0}^{r} k\\[-2pt]
    &= (n-1) \cdot (r+1) - \frac{r \cdot (r+1)}{2}\\[-4pt]
    &= -\frac{1}{2}r^2 + r \left(n- \frac{3}{2}\right) + (n-1)
    \label{eq:sum-r}
\end{align}
For a given row $r$, we have $\text{idx} \in [S_{r-1}, S_r)$. Thus we require $S_{r-1} \overset{!}{=} \text{idx}$ and solve for $r$. After some simple algebraic manipulations, we obtain:
\begin{align}
    0 &= -\frac{1}{2} r^2 + r \underbrace{\left(n - \frac{1}{2}\right)}_z - \text{idx}\\
    \iff 0 &= r^2 + r (-2z) + 2 \cdot \text{idx}\\
    \implies r(\text{idx}) &= z - \sqrt{z^2 - 2\cdot \text{idx}}
    \label{eq:row-index}
\end{align}
We only use the negative branch of the square root since we want the row to \textit{increase} with increasing indices. Furthermore, we require the row index to be an integer, so we round down $\lfloor r(\text{idx}) \rfloor$, such that the row stays the same for a range of indices (those that refer to the same row). Column $c$ is then given by:
\begin{align}
    c(\text{idx}) &\coloneqq r + 1 + (\text{idx} - S_{r-1})\\
    &= \frac{1}{2} r^2 + r \left(\frac{3}{2} - n\right) + (\text{idx} + 1)
    \label{eq:col-index}
\end{align}
Finally, note that $S_{n-1}$ conveniently gives us the number of elements in the upper triangular part of the adjacency matrix, \ie the number of edges in our graph. With \eqref{eq:sum-r}, we find:
\begin{align}
    \small{\text{num edges}}
    &= S_{n-1} = \frac{1}{2} n^2 - \frac{1}{2} n
    = \frac{n(n-1)}{2}
    \label{eq:num-edges}
\end{align}

With \eqref{eq:row-index} and \eqref{eq:col-index}, we determine the respective row and column in the adjacency matrix for a given index \eqref{eq:idx}. We use 64-bit floating point numbers to mitigate precision errors during row and column calculations on big graphs.

\textbf{Memory.} For every thread, one could allocate an array of arrays on the heap inside \textit{global} GPU memory to store the score matrix. However, this is very costly and immensely slows down the kernel execution. Instead, we rely on \textit{shared memory} for all threads in a block and let them share one big score matrix as to minimize the number of allocations needed. However, the score matrix dimensions differ in size depending on the word lengths of $A$ and $B$. Our solution is to look up the \underline{maximum word length $q$} before the kernel is launched, and allocate enough shared memory, \ie for every thread, we assume the worst-case scenario of a $(q+1)\times(q+1)$ matrix and allocate $(q+1)^2 \cdot \text{blockDim.x}$ chars (\tcboxverb{i8}) as linear shared memory. Then, inside each thread, we retrieve a pointer to the start of the memory reserved for this thread by adding $\text{threadIdx.x} \cdot (q+1)^2$ to the base pointer. To finally access element $(i,j)$ in the score matrix, we calculate $i\cdot \bigl(\lenn(B) + 1\bigr) + j$. This may not use all allocated memory, but it is a trade-off between memory usage and performance; with our approach we avoid bank conflicts.

\textbf{Block dimension.} This dimension specifies the number of threads inside a block and is limited to 1024. We want to maximize shared memory available per block in order to minimize the number of memory allocations. Before kernel launch, for each $u\in [1,1024]$, the shared memory size for one block ($u\cdot (q+1)^2$~bytes) is computed. The greatest $u$ is chosen such that the maximum available shared memory per block is not exceeded.

\section{Evaluation}
\label{sec:eval}

We deploy our GPU code on a consumer Nvidia GeForce GTX 1060\footnote{We use the Driver Version 572.42 and CUDA Toolkit 12.8 inside WSL2 (Ubuntu 22.04 jammy).} with 6GB GDDR5. Code changes are verified by comparing the resulting binary edge weights file with the one generated by our parallelized Rust implementation on the CPU. This baseline helps to identify errors, which could otherwise remain unnoticed. During our tests, we define a manual threshold to cap the number of words. We sort them according to their frequency as we are interested in relationships between the most commonly used words.

\vspace{-0.2em}

\textbf{Performance.} A fair comparison between the CPU and GPU implementation is not possible since focus was put in optimizing the GPU code. To give an order of magnitude, the parallelized Rust CPU implementation (without the subsequent sorting) takes around $\qty{12}{\s}$ (for 10,000 nodes), \qty{42}{\s} (for 20,000 nodes) and \qty{93}{\s} for 30,000 nodes on a 4-core Intel i7-6700 CPU. The implementation is limited to around 35,000 words when $\approx\qty{20}{\giga\byte}$ of RAM are available.

\begin{figure}[H]
    \centering
    \includegraphics[width=0.95\linewidth]{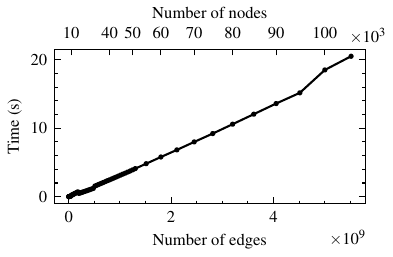}
    \caption{Performance of the GPU code for different number of nodes $n$. Number of edges via \eqref{eq:num-edges}.}
    \label{fig:timing}
\end{figure}

To test the performance of the GPU code, we measure the kernel execution time (including copying the results back to the host) for a range of number of nodes $n$ in the graph. For every $n$, we measure the duration 12 times\footnote{After every run, the device is re-initialized. Furthermore, we wait \qty{2}{\s} after every run before a new one starts.} and calculate mean and variance. The results are depicted in \autoref{fig:timing}. The variance is not shown as it is too small to be visible (always less than \qty{1}{\ms}). For $20,000$ words, the GPU code takes $\qty{484}{\ms}$ on average, while the CPU implementation needs $\qty{42}{\s}$. Up to $100,000$ nodes (\ie up to almost 5 billion edges), the GPU implementation takes less than \qty{20}{\s}. \autoref{fig:timing} also reveals the linear trend of time with increasing number of edges, which was to be expected since a thread is launched for every edge.

\vspace{-0.37cm}

\vspace{-0.5cm}

Our GPU implementation is limited by the global memory (\qty{6}{\giga\byte} for the GPU at hand). This memory is used to store the resulting edge weights, \ie one byte per edge. The maximum number of edges we can handle is therefore the available memory divided by 1~byte. To obtain the corresponding number of nodes, we solve \eqref{eq:num-edges} for~$n$:
\begin{align}
    n = \frac{1}{2} + \sqrt{\frac{1}{4} + 2 \cdot \text{num edges}}
\end{align}
On the GPU at hand, we can handle up to around 107,000 nodes (mean time $\approx \qty{21.3}{\s}$) before experiencing \q{CUDA out of memory errors}. Currently we detect the limit, but do not implement a mechanism to go beyond it. One way could be to copy the results back to the host and continue the computation while shifting the index back to $0$. The results are then concatenated on the host. For the further evaluation, we shall content ourselves with the results for the first 100,000 words, which already contain a wealth of information of the French language\footnote{For 100,000 words, the binary file that only holds the edge weights in row-major order, is \qty{4.66}{\giga\byte} in size.}.

\begin{figure}[H]
    \centering
    \includegraphics[width=0.9\linewidth]{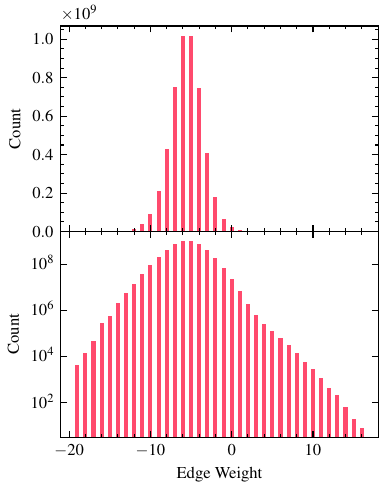}
    \caption{Histogram of edge weights (linear and logarithmic scale). Mean: $-1.5$, range: $[-19, 16]$.}
    \label{fig:hist}
\end{figure}

\begin{figure}[H]
    \centering
    \includegraphics[width=0.9\linewidth]{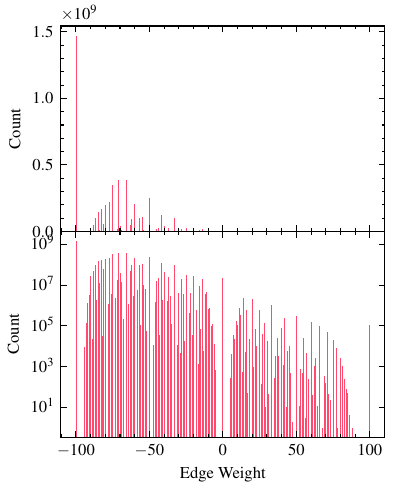}
    \caption{Histogram of normalized edge weights. Linear and logarithmic scale.}
    \label{fig:hist-norm}
\end{figure}

\textbf{Graph application.} \autoref{fig:hist} shows the histogram of edge weights. It strongly resembles a normal distribution, which is probably due to how word lengths are distributed in the language. Note that the minimum and maximum achievable alignment score for a word pair depends on the word lengths. To efface this dependency, we normalize every score by dividing by $\max(\text{len}(A), \text{len}(B))$ and then multiply by 100. The resulting histogram is shown in \autoref{fig:hist-norm}. It does not resemble a normal distribution anymore. Most nodes have the smallest score of $-100$. There is a gap ($\approx$ width~10) around edge weight~0 that we cannot explain. The global trend is that many edges have a strong negative edge weight, while a smaller proportion has a strongly positive one.\label{paragraph:normalization}

\vspace{-1.1em}

\vspace{-1.2em}

\vspace{-1.4em}

We visualize the resulting graph in the open-source software \href{https://gephi.org/}{Gephi} \cite{tools_gephi}. As Gephi is far from being able to display 5 billion edges at the same time (let alone import such a file), we have to select specific ranges of edge weights we are interested in. We focus on the positive edge weights as they indicate a higher similarity between words.

Gephi implements various force-directed graph drawing algorithms that can help us gain a better understanding of the data. The principle of these algorithms is that nodes repulse each other, while edges act as springs to pull connected nodes together (taking into account the edge weights). Here, we exclusively use the algorithm by Yifan Hu \cite{yifanhu} as we found it to be the most efficient and reliable for our data. Furthermore, we run a modularity analysis using the Louvain method \cite{louvain} implemented in Gephi. This method is used to detect communities in the graph, \ie groups of nodes that are more connected to each other than to the rest of the graph. We color the nodes according to the community they belong to\footnote{Colors don't match between \textit{different} graphs in the following.}.

\begin{figure*}[t]
    \centering
    \includegraphics[width=0.96\textwidth, trim=0cm 5.5cm 0cm 5.5cm, clip]{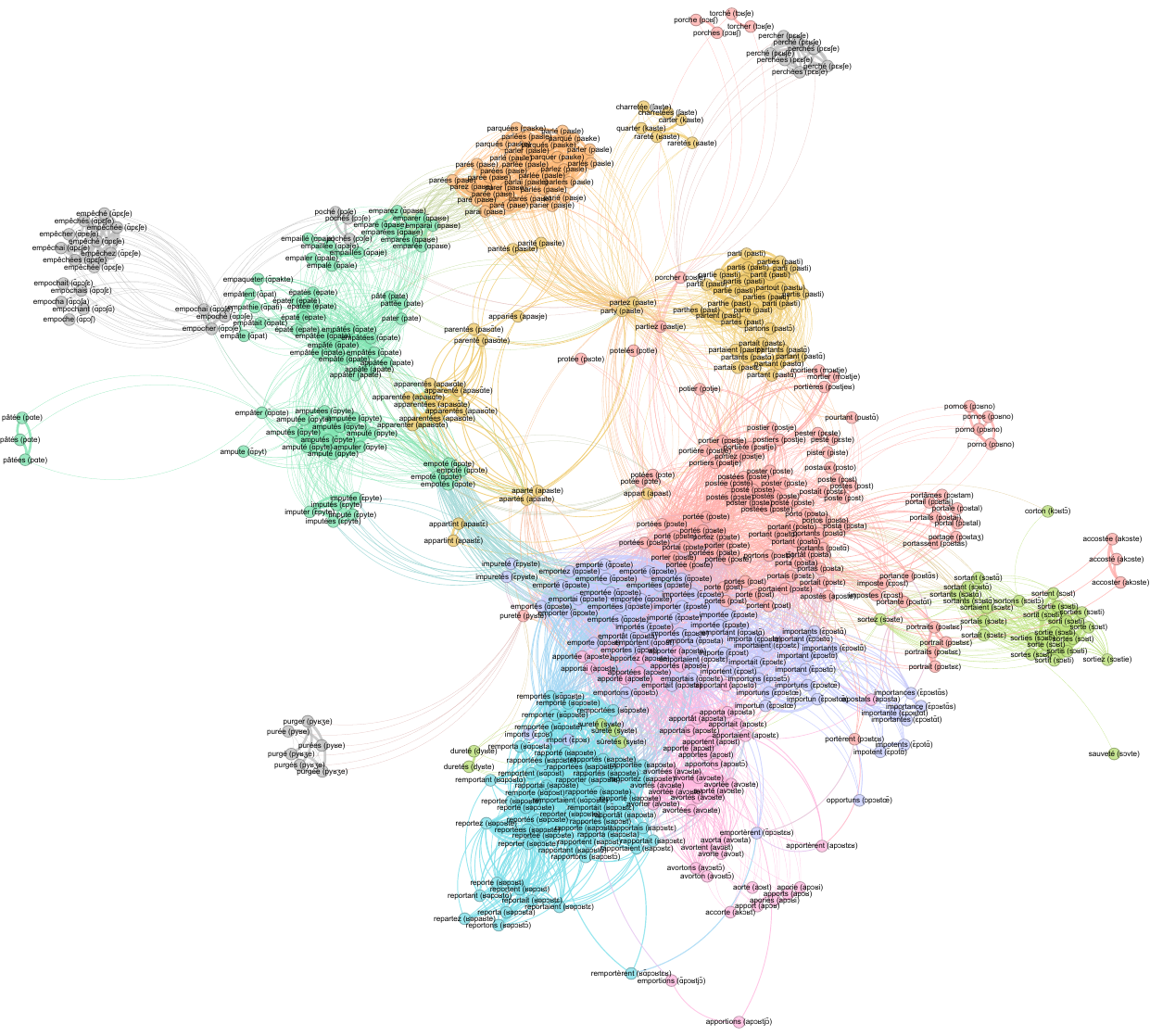}
    \caption{Ego-network (depth 3) of the word \textit{emporter} (to take away) for edge weights in the range $[40,49]$. This subgraph contains 449 nodes and 5921 edges.}
    \label{fig:emporter-ego}
\end{figure*}

\begin{figure}[H]
    \centering
    \includegraphics[width=\linewidth, trim=1cm 3.4cm 0.5cm 3.2cm, clip]{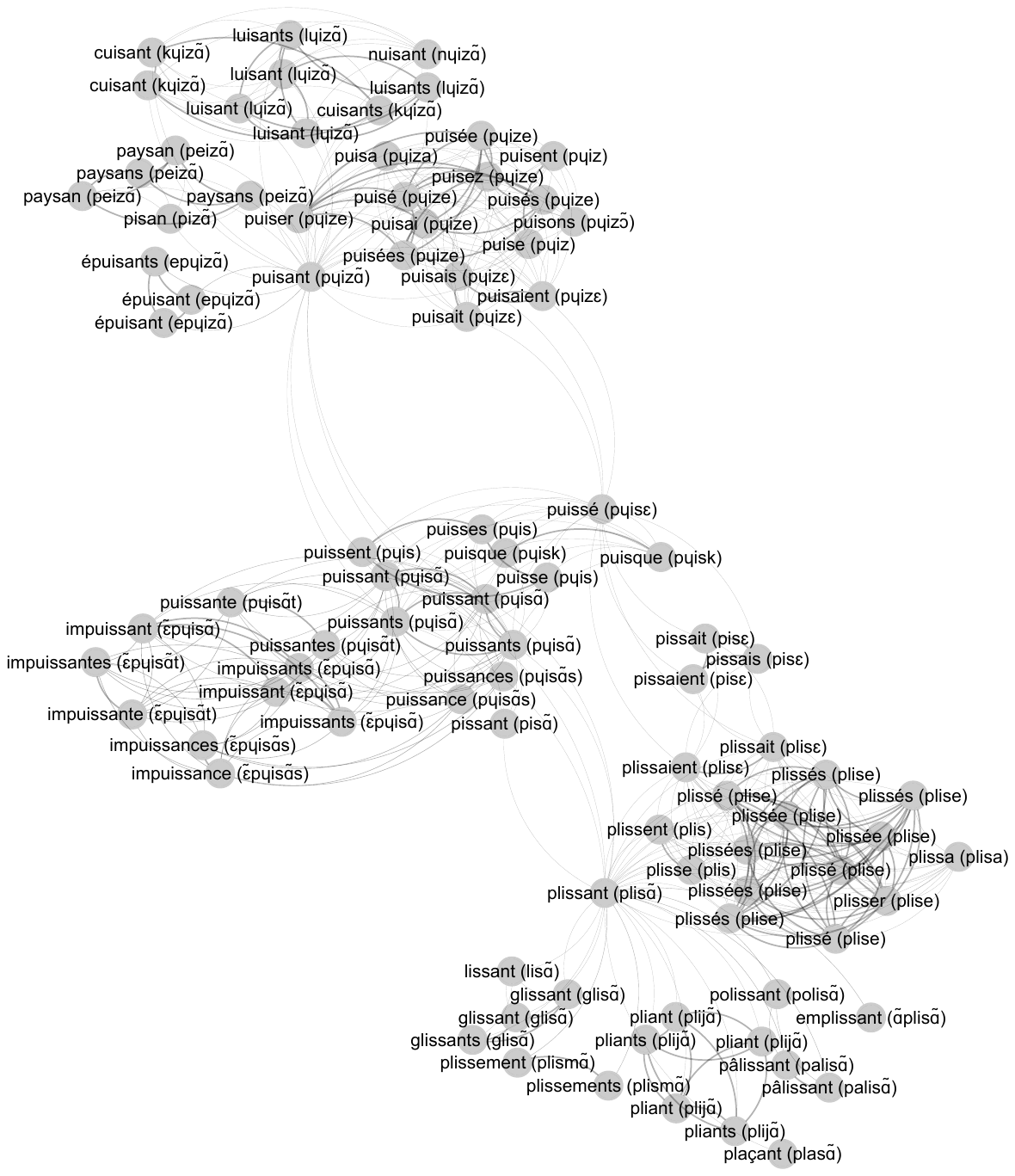}
    \caption{Ego-network (depth 3) of the word \textit{puissant} (powerful) for edge weights $\in [60, 100]$.}
    \label{fig:puissant-ego}
\end{figure}

\begin{figure}[H]
    \centering
    \includegraphics[width=\linewidth, trim=1cm 4.8cm 0.5cm 3.2cm, clip]{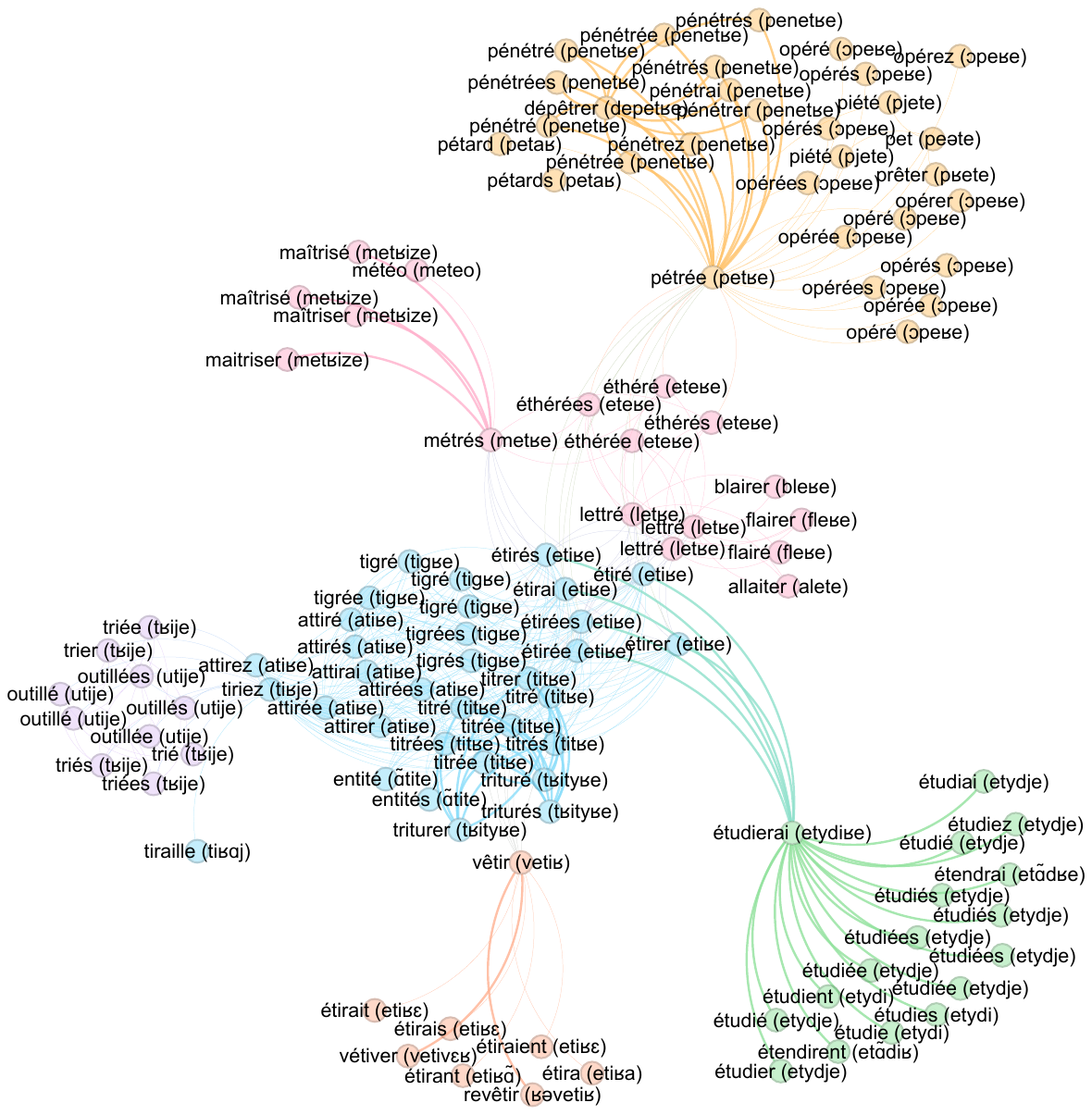}
    \vspace{0.14cm}
    \caption{Ego-network (depth 3) of the word \textit{étirer} (to stretch) for edge weights $\in [40,49]$.}
    \label{fig:etirer-ego}
\end{figure}

Some resulting ego-networks are shown in figures \ref{fig:emporter-ego}, \ref{fig:puissant-ego} and \ref{fig:etirer-ego}. These graphs are constructed by locating neighbors of a word, then finding the neighbors of these neighbors up to a depth of~3. Different slices of edge weights are used, \eg in \autoref{fig:emporter-ego}, we have filtered the graph for edge weights in the range $[40,49]$ beforehand; they account for a total of 458,529 edges. The networks clearly show that our implementation is working correctly and that Needleman–Wunsch indeed gives a meaningful metric in this context:

\begin{itemize}[leftmargin=0.3cm]
    \item Words with the same pronunciation are as close as possible to each other and reside in the same community. In \autoref{fig:etirer-ego}, the words \textit{étudier}, \textit{étudié}, \textit{étudiée}, \textit{étudiées}, \textit{étudiez}, etc. are close to each other. In \autoref{fig:emporter-ego}, inside the red group in the middle-right part, we find different adjective endings for the word \textit{porté} (corresponding to gender and number of the noun it describes, \ie adjective agreement): \textit{porté}, \textit{portée}, \textit{portés}, \textit{portées}. We also find the words \textit{porter} and \textit{portez} here with the same prononciation as \textit{porté}. Note that it is claimed that \textit{portai} has the exact same prononciation, which is not correct (the last phoneme is different).
    
    \item Of greater interest are similar sounding words; as hoped for, they are close to each other in the graph. For example, in \autoref{fig:emporter-ego}, we find edges like \textit{emporté} -- \textit{porter}, \textit{emporté} -- \textit{importé}, \textit{importé} -- \textit{impureté} and \textit{impureté -- pureté}. It is also remarkable that words in one group almost exclusively start with the same letter and correspond to the same word root: \textit{sortir} in the green group on the right, \textit{porter} and \textit{poster} in the red group, \textit{emporter} and \textit{importer} in the violet group, \textit{apporter} in the rose group, \textit{rapporter}, \textit{reporter} and \textit{remporter} in the blue group etc.
    
    \item Words are connected to each other to varying degrees. In \autoref{fig:puissant-ego}, the edge between \textit{puisant}\footnote{Note that this is not missing an additional \textit{s}. This is the word in a phrase like \q{en \textit{puisant} de l'eau, ...}.} -- \textit{épuisant} has weight $\nicefrac{200}{3}\approx 66.7$ (after normalization), while that of \textit{puisant} -- \textit{paysans} has the smaller weight $\nicefrac{300}{5} = 60$.
\end{itemize}

\begin{table}[H]
    \centering
    \begin{tabular}{l*{6}{>{\centering\arraybackslash}p{0.2cm}}}
        \toprule
        \textit{paysans}
        & & \textipa{p} & \textipa{e} & \textipa{i} & \textipa{z} & \textipa{A}\\
        \midrule
        \textit{puisant}
        & & \textipa{p} & \textipa{\textturnh} & \textipa{i} & \textipa{z} & \textipa{\~A}\\
        \midrule
        \textit{épuisant}
        & \textipa{e} & \textipa{p} & \textipa{\textturnh} & \textipa{i} & \textipa{z} & \textipa{\~A}\\
        \bottomrule
    \end{tabular}
    \caption{Optimal alignment of three words.}
    \label{tab:align-puisant}
\end{table}

\vspace{-1.5em}

\autoref{tab:align-puisant} shows the corresponding optimal alignments. With a match score of $1$, a mismatch score of $-1$ and a gap penalty $p=-1$, we find a score of~3 (\textit{puisant} -- \textit{paysans}) and 4 (\textit{puisant} -- \textit{épuisant}). With the normalization discussed on \autopageref{paragraph:normalization}, we indeed obtain the aforementioned edge weights. This example also illustrates that fine-tuning match/mismatch score and gap penalty is important to obtain meaningful results. One might want a greater distance between \textit{paysans} and \textit{puisant} since the \textipa{/u/} is replaced by the very different sounding \textipa{/a/}. The match/mismatch scores can be adjusted for every pair of phonemes to reflect the subtleties of a language by changing the coefficients in the similarity matrix (see \autoref{algstep:sim} of \autoref{alg:needleman-wunsch}).

\vspace{-0.8em}

\begin{figure}[H]
    \centering
    \includegraphics[width=\linewidth, trim=1cm 4.8cm 0.5cm 3.2cm, clip]{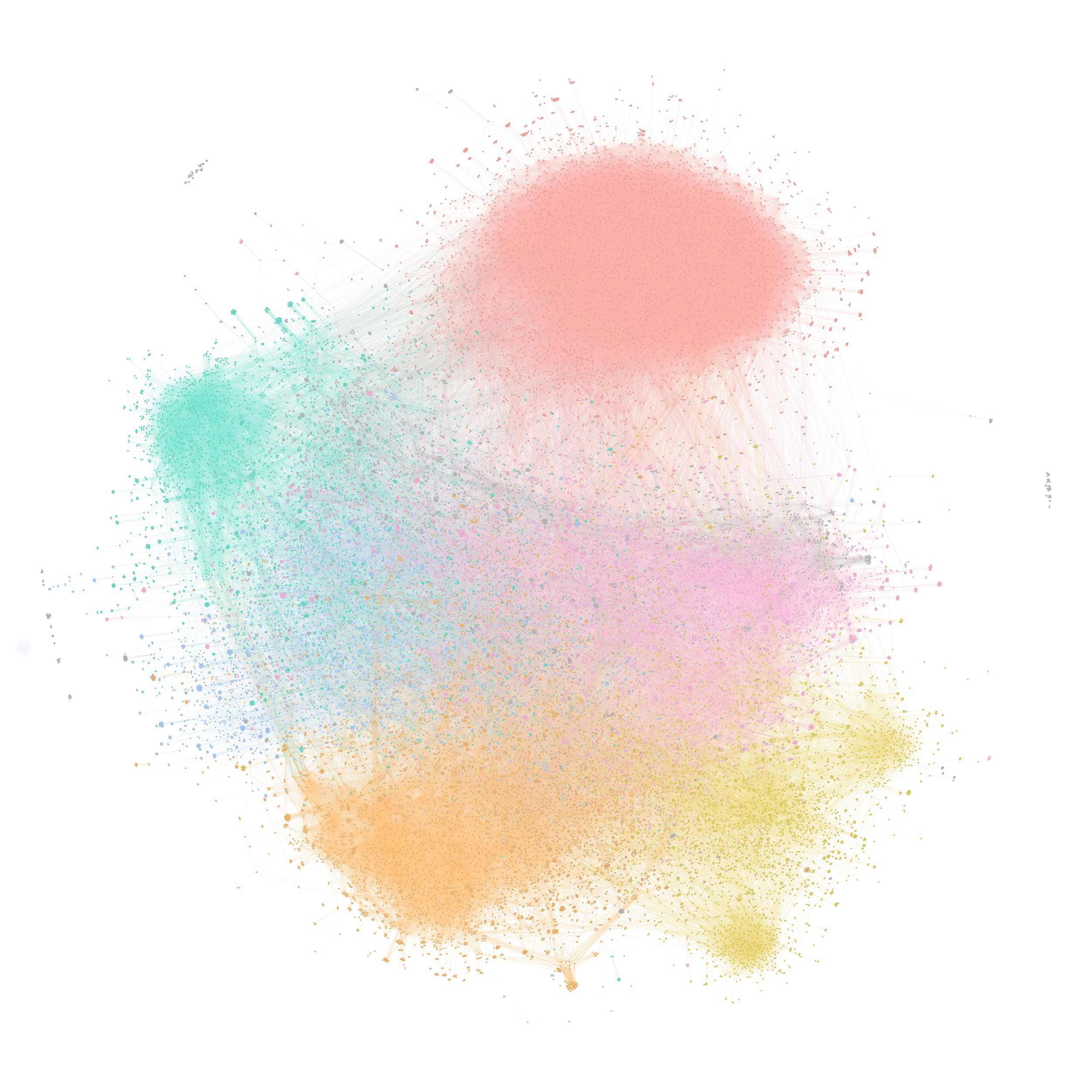}
    \caption{Overview of the graph containing edge weights in the range $[8,10]$. It consists of 30,185 nodes and 306,473 edges.}
    \label{fig:big-view}
\end{figure}

\vspace{-1.2em}

To broaden the view, we consider a bigger graph in \autoref{fig:big-view}, with normalized edge weights 8, 9 or 10. Communities found with Louvain are again color-coded. This very tiny slice of edge weights (compare \autoref{fig:hist-norm}) can already reveal interesting partitions of a language.

\begin{itemize}
    \item The red community at the top consists almost exclusively of words ending in \textipa{/j\~O/}, \eg \textit{illustration}, \textit{proportions}, \textit{réalisation}, \textit{fluctuation}, \textit{documentation}, \textit{émancipation}, \textit{articulation} etc.
    
    \item Below, the rose group encompasses words that end with \textipa{/\OE K/}, \eg \textit{aspirateur}, \textit{consommateur}, \textit{illustrateur}, \textit{radiateur}. However, this group is less homogeneous and also contains words ending in \textipa{/ktiv/}, \eg \textit{interactives}, \textit{réparatrice}, \textit{instructif}, \textit{instinctives}. 
    
    \item The orange group on the bottom mainly comprises words ending in \textit{e}: \textit{personnalité}, \textit{caractérisé}, \textit{centraliser}, \textit{hospitalité}, \textit{réalité}, \textit{vivacité}, \textit{patienter}, \textit{spécialiser}, \textit{sensibiliser}, \textit{modéliser}.
    
    \item In the cyan group on the far left, we find many words that end in \textipa{El}, \eg \textit{personnel}, \textit{conventionnel}, \textit{professionnel}, \textit{conditionnelles}, \textit{exceptionnelle}, \textit{émotionnels}, \textit{sensationnels}. However, this group is harder to describe since it also includes different words like \textit{occasionnant}, \textit{collectionneur}, \textit{frictionne}, \textit{crayonné}, \textit{positionnement}, where similar features are not obvious.
    
    \item The yellow group on the bottom right is easier to classify: it contains words ending in \textipa{Zi} or \textipa{Zik}. Among others: \textit{méthodologique}, \textit{topologique} and \textit{topologie}, \textit{holistique}, \textit{diplomatique}, \textit{dramaturgie}, \textit{cytologie}, \textit{phénoménologie}, \textit{biologie}, \textit{écologique}, \textit{synchronique}. 
\end{itemize}

We also tried to identify common groups of words located \textit{in-between} two communities, in the expectation of finding a \q{morphing} behavior between them. However, this could not be observed in the graphs and we were not able to find a meaningful way to group these words.

Finally, having translated the problem into a graph structure also allows us to use graph algorithms to discover interesting properties. As an example, Gephi implements the \textit{shortest path algorithm}: users can click on two words to display the shortest path that link them (in a graph filtered for the strongest edges). With this, we can find chains like the following (read them aloud to hear the phonetic similarity):
\begin{itemize}
    \item trottoir $\rightarrow$ entrevoir $\rightarrow$ devoir $\rightarrow$ voire $\rightarrow$ voile $\rightarrow$ val $\rightarrow$ valait $\rightarrow$ fallait $\rightarrow$ falaise
    \item falaise $\rightarrow$ fallait $\rightarrow$ palais $\rightarrow$ passais $\rightarrow$ dépassait $\rightarrow$ dépendait $\rightarrow$ répondait $\rightarrow$ répond $\rightarrow$ raison $\rightarrow$ maison
    \item confusion $\rightarrow$ conclusion $\rightarrow$ exclusion $\rightarrow$ explosion $\rightarrow$ exposition $\rightarrow$ explications $\rightarrow$ respiration $\rightarrow$ précipitation $\rightarrow$ présentation $\rightarrow$ présenta $\rightarrow$ présente $\rightarrow$ présence $\rightarrow$ présidence $\rightarrow$ résidence $\rightarrow$ résistance $\rightarrow$ existence
\end{itemize}

\vfill\null

\pagebreak

\section{Conclusion \& Outlook}
\label{sec:conclusion}

We presented a method to calculate similarity between words based on their phonetic \gls{ipa} transcription, \ie how they sound when spoken. For this task, the Needleman–Wunsch algorithm with similarity matrix and gap penalty was employed. We implemented this algorithm in Rust and parallelized it on a CPU using the \textit{Rayon} library and on a consumer Nvidia GPU using the CUDA framework with the \textit{cudarc} library, while writing the kernel itself in \Cpp. We detailed choice of data structures, indexing and memory layout to optimize the performance. For a graph with 100,000 nodes (words) and almost 5 billion edges (word-pairs), the algorithm takes less than \qty{20}{\s} (including copying back to host). Consistency between the CPU and GPU implementations was verified up to 30,000 nodes\footnote{After that, the CPU implementation becomes too slow and consumes too much memory.}.

Future work could include an adapted version that can read the whole graph of more than 600,000 nodes at the same time by copying back intermediate results from the GPU to the CPU as outlined in the evaluation. To further speed up the computation, one could consider a more fine-grained parallelization at the level of the score matrix calculation. This is not trivial due to the dependencies between the cells of the matrix. A stencil computation approach might be feasible.

We also demonstrated the practical usability of the resulting edge weights by examining the graphs in Gephi. Even just by looking at small subsections of the full graph, we can deduce interesting properties of the language, \eg community detection revealed groups with similar word endings and groups with the same root but different endings. Constructing ego-networks allows to find words that are phonetically similar to a given word. One can envision an online platform where this functionality is offered to users to find rhymes or similar sounding words. This can be helpful for language learners as well to foster the playful exploration of a language. Beyond that, future work might tackle the following points:

\begin{itemize}
    \item Fine-tune the similarity matrix for language subtleties. In this document, we always used a fixed match/mismatch score of $1$/$-1$, while in reality, some phonetic substitutions sound more similar than others. For example, replacing a vowel by a consonant might be more severe than replacing a vowel by another vowel. One might also want to experiment with the gap penalty that was set to $-1$ throughout.
    
    \item So far, we considered a dataset of the French language, while the method is not limited to it. By simply replacing the dataset with one of another language, we can apply the method of this paper to any language that has phonetic transcriptions available. This opens the possibility to compare the phonetic structure of different languages and find similarities between them.
    
    \item The Louvain algorithm \cite{louvain} for community detection inherently constructs a multi-level hierarchy of communities. Looking at a lower granularity level (more high-level view) could reveal interesting results. For this purpose, we should consider a parallel Louvain implementation that can work on the whole graph. Limiting oneself to slices of edge weights (as done here) might bias results or leave interesting structures unnoticed.
    
    \item The size of the dataset can be reduced by merging words with the same lexeme, \eg for \textit{remportés}, \textit{remportant}, \textit{remportée}, \textit{remportées} etc. we could only consider the lexeme \textit{remporter}. This could help gain a better and more general understanding of the phonetic structure of the language as the graphs would not be cluttered with variations of the same word.
\end{itemize}

\section*{Acknowledgements}

I would like to thank Christophe Picard for his lecture that introduced me to GPU computing and for reviewing this paper. Furthermore, I am grateful to the Gephi contributors for their software that made it possible to visualize the graphs in this paper. I also want to thank all people who compiled the various datasets, without which this work would not have been possible.

\renewcommand*{\glsgroupskip}{} 
\printglossary[type=\acronymtype]
\end{multicols*}

\printbibliography[
    heading=bibintoc,
    title={Bibliography},
    keyword={lit}
]
\printbibliography[
    title={Data sources \& Tools},
    keyword={data}
]

\end{document}